\documentclass[10pt,twocolumn,letterpaper]{article}

\usepackage{iccv}
\usepackage{times}
\usepackage{epsfig}
\usepackage{graphicx}
\usepackage{amsmath}
\usepackage{amssymb}
\usepackage{bm}
\usepackage{booktabs}
\usepackage{multirow}
\usepackage{subfigure}
\usepackage{enumitem}
\usepackage{caption}
\usepackage{bbm}
\usepackage{colortbl}
\usepackage{url}
\usepackage{tabularx}
\usepackage{lscape}
\usepackage{xcolor}

\usepackage[breaklinks=true,bookmarks=false]{hyperref}

\iccvfinalcopy 

\def\iccvPaperID{****} 

\ificcvfinal\fi

\begin{document}

\title{SimFIR: A Simple Framework for Fisheye Image Rectification with Self-supervised Representation Learning}

\author{Hao~Feng$^{1,2}$~~~~~Wendi~Wang$^{1}$~~~~~Jiajun Deng$^{3}$~~~~~Wengang~Zhou$^{1,4,}$\thanks{Corresponding authors: Wengang Zhou and Houqiang Li}~~~~~Li~Li$^{1}$~~~~~Houqiang~Li$^{1,4,}$\footnotemark[1] \\
	{\normalsize $^{1}$ CAS Key Laboratory of Technology in GIPAS, EEIS Department, University of Science and Technology of China} \\
  	{\normalsize $^{2}$ Zhangjiang Laboratory, Shanghai, China} \quad 
 	{\normalsize $^{3}$ The University of Sydney} \\
	{\normalsize $^{4}$ Institute of Artificial Intelligence, Hefei Comprehensive National Science Center} \\
	{\tt\small \{haof,wendiwang\}@mail.ustc.edu.cn, jiajun.deng@sydney.edu.au, \{zhwg,lil1,lihq\}@ustc.edu.cn}
}


\maketitle
\ificcvfinal\thispagestyle{empty}\fi

\begin{abstract}
In fisheye images, rich distinct distortion patterns are regularly distributed in the image plane. These distortion patterns are independent of the visual content and provide informative cues for rectification. 
To make the best of such rectification cues,
we introduce SimFIR, a simple framework for fisheye image rectification based on self-supervised representation learning.
Technically, we first split a fisheye image into multiple patches and extract their representations with a Vision Transformer (ViT).
To learn fine-grained distortion representations,
we then associate different image patches with their specific distortion patterns based on the fisheye model, and further subtly design an innovative unified distortion-aware pretext task for their learning.
The transfer performance on the downstream rectification task is remarkably boosted, which verifies the effectiveness of the learned representations.
Extensive experiments are conducted,
and the quantitative and qualitative results demonstrate the superiority of our method over the state-of-the-art algorithms as well as its strong generalization ability on real-world fisheye images.
\end{abstract}

\section{Introduction}
Fisheye lenses have been widely used in systems for video surveillance~\cite{muhammad2018efficient, lin2012integrating,ferryman2000visual}, autonomous driving~\cite{grigorescu2020survey,6248074} and robotic applications~\cite{courbon2007generic,courbon2012evaluation}, 
thanks to their extremely wide field of view (FoV).
However, images captured by the fisheye lenses are inevitably distorted, which are not directly applicable for the downstream analysis, such as object detection~\cite{plaut20213d,goodarzi2019optimization}, pose estimation~\cite{zhang2021automatic,aso2021portable,rhodin2016egocap} and scene segmentation~\cite{kumar2021syndistnet,deng2017cnn,blott2018semantic}.
Over the past few years,
fisheye image rectification has become an emerging research topic.

\begin{figure}[t]
	\centering
	\includegraphics[width=0.96\columnwidth]{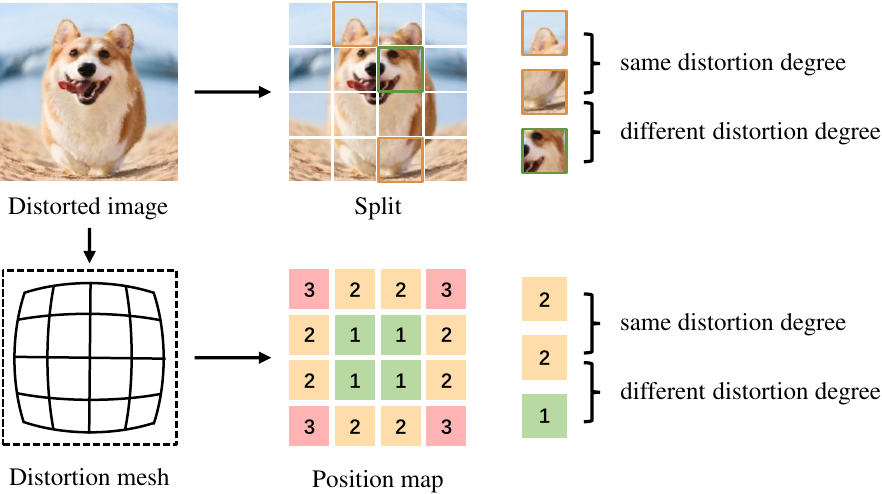}
	\caption{An illustration of fisheye distortions. The distortions in fisheye images are radially distributed and centrally symmetrical. The distortion degree increases with distance from the optical axis/image center.}
	\label{fig:motivation}
\end{figure}

In the literature, some classical methods~\cite{barreto2005fundamental,kannala2006generic, hartley2007parameter, kukelova2011minimal, puig2011calibration,grossberg2001general,fan2022wide} estimate epipolar geometry with images from different viewpoints.
However, multiview images are not always accessible, which limits the real applications of these methods.
To overcome this limitation,
other classical methods are based on the detection of plumb lines~\cite{thormahlen2003robust,wang2009simple,bukhari2013automatic,devernay2001straight,zhang2015line,barreto2009automatic}.
Nevertheless, taking the detected plumb lines as the basis of fisheye image rectification inevitably causes error accumulation.
Recently, deep learning based solutions~\cite{rong2016radial,bogdan2018deepcalib,yin2018fisheyerecnet,liao2019dr,li2019blind,yang2021progressively,liao2021multi,liao2020model,xue2019learning} have become a promising alternative to the traditional methods. There are two directions among deep learning based methods. One direction~\cite{rong2016radial,bogdan2018deepcalib,yin2018fisheyerecnet,xue2019learning} attempts to predict the distortion parameters to infer pixel-wise displacement for rectification. 
The other direction~\cite{liao2019dr,yang2021progressively,liao2021multi,liao2020model} directly predicts the rectified images with specifically designed architectures.
Although these methods are reported with state-of-the-art results,
the intrinsic distortion characteristics of fisheye images are largely ignored, described next. 

Generally, as illustrated in Fig.~\ref{fig:motivation}, the distortions in fisheye images are radially distributed and centrally symmetrical.
The distortion degree increases with distance from the image center along the radial axis.
It should be noted that here we follow the existing state-of-the-art rectification algorithms~\cite{rong2016radial,bogdan2018deepcalib,yin2018fisheyerecnet,liao2019dr,li2019blind,yang2021progressively,liao2021multi,liao2020model,xue2019learning} to only consider the ideal fisheye images, in which the distortion center locate at the image center.
Furthermore, 
when dividing a fisheye image into patches, the distortion patterns and degrees vary among patches at different distances from the image center. 
To improve fisheye image rectification, it is beneficial to model these fine-grained distortion patterns that provide cues for the rectification.

Formally, we present SimFIR, a new framework for fisheye image rectification based on self-supervised representation learning~\cite{yao2023towards}. SimFIR introduces an effective way for encoding the fine-grained distortion patterns in fisheye images, by learning the associations between the local image region and their distinct distortion pattern.
Specifically, we first split a fisheye image into multiple patches and label them with a position map (see Fig.~\ref{fig:motivation}) based on their specific distortion degrees.
Then, these image patches are embedded as input tokens of Vision Transformer (ViT)~\cite{dosovitskiy2020image} to abstract patch representations.
Given these patch representations, we introduce a unified distortion-aware pretext task to optimize the network.
On the one hand, we perform contrastive learning with InfoNCE loss~\cite{oord2018representation} to narrow the representation of patches with the same distortion pattern while enlarging the feature distance with others.
On the other hand, 
we devise a classification supervision to predict the patch-wise quantified distortion degree, aiming to differentiate the distinct local distortion patterns.
After pre-training, the network is capable of extracting the fine-grained distortion representations, regardless of the image textures.
Thereafter, the learned distortion representations are transferred to the downstream rectification task.

Besides, we introduce a new pipeline for fisheye image rectification, which unwarps fisheye images with a dense flow field, $i.e.,$ the warping flow, aiming to process images of arbitrary resolution without complex post-processing~\cite{li2019blind}.
Then, to facilitate training, 
we construct a well-annotated synthesized dataset that covers various scenes and distortion levels.
To validate the learned representations,
we conduct extensive experiments by pre-training the network with our self-supervised paradigm and then transferring it to the downstream rectification task.
The results demonstrate the effectiveness of our method as well
as its superiority over existing state-of-the-art algorithms.
In addition, we show the strong generalization ability of our method on processing real-world fisheye images.

In summary, we make three-fold contributions:
\vspace{-0.01in}
\begin{itemize}
    \item 
    We present SimFIR, a novel self-supervised based framework, to improve fisheye image rectification 
    by extracting its fine-grained distortion patterns.
    \item 
    We introduce a new pipeline for fisheye image rectification that unwarps images with backward warping flow, aiming to process images with any resolutions.
    \item
    We conduct extensive experiments to verify the merits of our method and report the state-of-the-art performance with our proposed method.
\end{itemize}

\section{Related Work}
Fisheye image rectification has been widely explored in the computer vision community.
In the following, we separately discuss the methods of each group.

\smallskip
\textbf{Rectification Based on Multi-view Calibration.}
Some traditional methods~\cite{barreto2005fundamental, hartley2007parameter, kukelova2011minimal, puig2011calibration, henrique2013radial, sturm2004generic,scaramuzza2006flexible} calibrate fisheye images by finding the corresponding feature points from multi perspectives.
Typically, FMC~\cite{barreto2005fundamental} proposes to estimate the epipolar geometry of two views with different distortion factors.
AMS~\cite{kukelova2011minimal} explores the minimal solution that calibrates radial distortion with only eight-point correspondences in two images.
However, such methods require special chessboards and multiple images are not always accessible, unfriendly to their practical applications.

\smallskip
\textbf{Rectification Based on Curve Line Detection.}
The overcome these limitations, 
line-based camera calibration methods~\cite{thormahlen2003robust,wang2009simple,bukhari2013automatic,devernay2001straight,barreto2005geometric,zhang2015line,chander2009summary} propose to utilize the prior that straight lines in the 3D world must project to straight lines in
the image plane.
Typically, 
SC~\cite{santana2016iterative} first detects the longest distorted lines within an image by applying Hough transformation~\cite{duda1972use} and then estimates the distortion parameters from these lines.
However, the curve detection is not robust and challenging, limiting their performance.

\begin{figure*}[t]
	\centering
	\includegraphics[width=1.82\columnwidth]{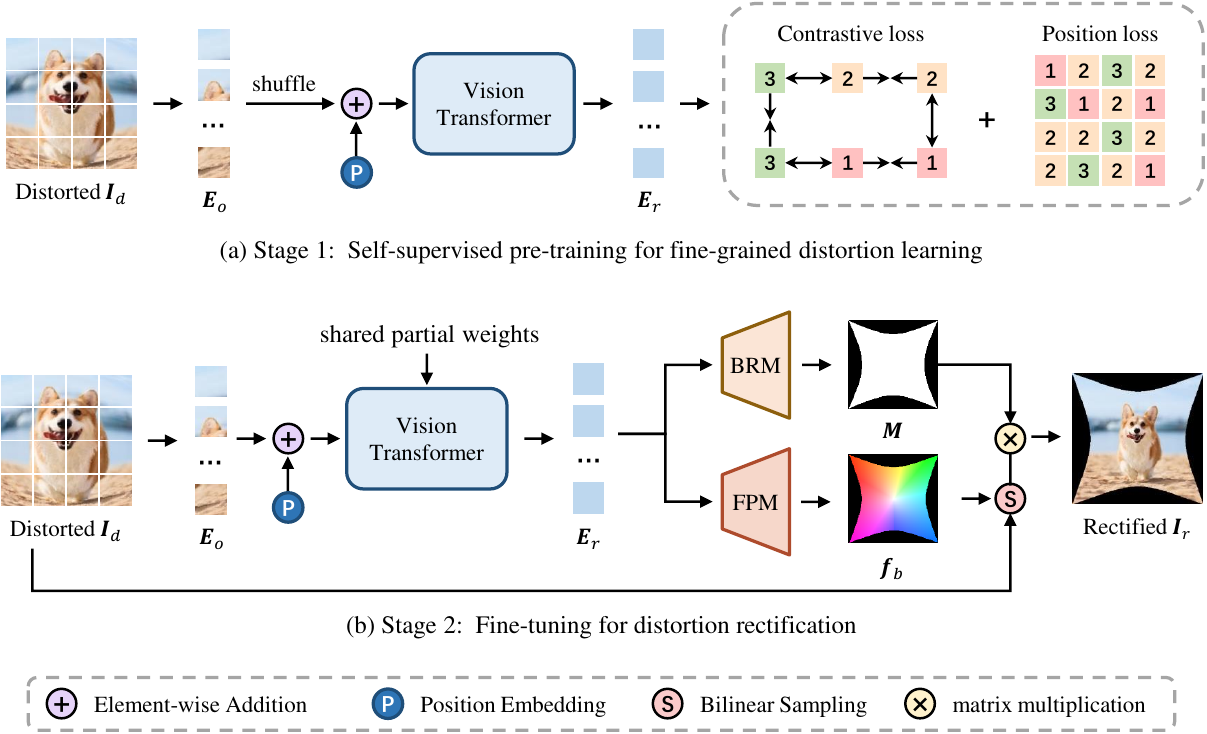}
	\caption{Framework of our method for fisheye image rectification. 
	It consists of two stages: 
	(a) self-supervised pre-training that learns the fine-grained distortion representation of fisheye image $\bm{I}_d$ with a unified distortion-aware pretext task;
	(b) fine-tuning for rectification which uses the learned representation to reconstruct the rectified image $\bm{I}_r$ with a flow field. ``BRM" and ``FPM" denote the proposed boundary refinement module and flow prediction module, respectively.
	}
	\label{framework}
\end{figure*}

\smallskip
\textbf{Rectification Based on Parameter Estimation.}
By predicting the distortion parameters of fisheye models,
some methods compute the per-pixel coordinate displacement to perform rectification. 
RLDC~\cite{rong2016radial} first employs a convolutional neural network to predict the distortion parameters.
DeepCalib~\cite{bogdan2018deepcalib} predicts the distortion parameters and the focal length with a network.
Given a single image, Blind~\cite{li2019blind} devises a unified framework to correct different types of geometric distortion by the estimation of the distortion parameter and forward warping flow~\cite{xie2020dewarping}. 
LCSL~\cite{xue2019learning} imposes an explicit geometry constraint onto the processes of the fisheye lens calibration and rectification.

\smallskip
\textbf{Rectification Based on Image Estimation.}
Methods following this direction commonly attempt to directly predict the rectified images.
DR-GAN~\cite{liao2019dr} exploits a generative adversarial network (GAN)~\cite{ledig2017photo} for rectified image synthesis, and proposes a low-to-high perceptual loss to capture the mapping relation between different structural images.
FE-GAN~\cite{chao2020self} learns pixel-level distortion flow from sets of fisheye
distorted images and distortion-free ones with the cross-rotation and intra-warping consistency.
SIR~\cite{fan2020sir} devises a self-supervised learning
idea for rectification by exploiting intra-model and
inter-model consistency.
DDM~\cite{liao2020model} proposes a distortion distribution map that intuitively indicates the global distortion features of a distorted image.
Recently,
MLC~\cite{liao2021multi} introduces a multi-level curriculum to train the rectification model hierarchically.
PCN~\cite{yang2021progressively} develops an encoder-decoder architecture to generate multi-scale rectified images progressively.


\section{Fisheye Model and Dataset}
We first introduce the distortion model for fisheye images. 
Since obtaining ground truth is difficult for real fisheye images,
we follow the common practices in the literature~\cite{yin2018fisheyerecnet,xue2019learning, bogdan2018deepcalib,liao2020model,liao2021multi,yang2021progressively}
to exploit synthetic data for training.
The polynomial model~\cite{kannala2006generic} of a fisheye image is formulated as follows,
\begin{equation}
    \theta_c=\sum_{i=1}^{N_k}{\lambda_i}{\theta_d^{2i-1}},N_k=1,2,3,4,\cdots,
\end{equation}
where $\theta_c$ denotes the angle of incident light, $\theta_d$ is the angle of the light after passing through the lens, $\lambda_i$ is the $i^{th}$ distortion parameter,
and $N_k$ is the number of distortion parameters.
According to the projection model for general pinhole cameras, we have $r_c = f \cdot tan\theta_c \approx f \cdot \theta_c$, where $f$ is the focal length of a camera, $r_c$ is the $L_2$ distance between the image center and an arbitrary point $P$ in a distortion-free image. Hence, 
the distortion model can be rewritten as,
\begin{equation}
    r_c=f\sum_{i=1}^{N_k}{\lambda_i}{\theta_d^{2i-1}},N_k=1,2,3,4,\cdots.
\end{equation}
Furthermore, based on the equidistant projection relation $r_d = f \cdot \theta _d$,
where $r_d$ is the $L_2$ distance between the image center and the mapping point $P'$ in the distorted image, we merge $f$ and $\theta_d$ to obtain the fisheye distortion model as,

\begin{equation}
    r_c=\sum_{i=1}^{N_k}{k_i}{r_d^{2i-1}},N_k=1,2,3,4,\cdots.
    \label{distortion_model}
\end{equation}
It depicts the radial distortion distribution of fisheye image and is used for the dataset generation (see Sec.~\ref{datasettext}).

\section{Methodology}
Our goal is to learn useful visual representations of fisheye images to improve their rectification.
Fig.~\ref{framework} presents the overview of our method, which consists of two stages: (a) self-supervised pre-training for distortion representation learning, and (b) fine-tuning for distortion rectification.
We elaborate on each stage in the following sections.

\subsection{Self-supervised Pre-training}
As shown in Eq.~\eqref{distortion_model}, fisheye images have an inherent property where the distortions are radially symmetrical, and the degree of distortion in a particular region is dependent on its position on the image plane. A greater radius leads to a higher degree of distortion. To enhance rectification by capturing these detailed distortion patterns, we propose a unified distortion-aware pretext task.

Specifically, on the one hand,
we use contrastive learning~\cite{chen2020simple} to narrow the patch features with the same distortion degree and push away the different ones.
This approach allows us to encode local distortion features while eliminating texture features. 
On the other hand,
we estimate the patch-wise quantified distortion degrees to learn fine-grained distortion features, by introducing a position loss (see Fig.~\ref{framework}).
To implement the idea,
we introduce a transformer-based model~\cite{vaswani2017attention}, described next.

\smallskip
\textbf{Patch Embedding.}
The standard transformer~\cite{vaswani2017attention} takes a sequence of token embeddings as input.
To adapt it, 
as shown in Fig.~\ref{framework}, given a fisheye image $\bm{I}_d \in\mathbb{R}^{H\times W\times 3}$, we first divide it into a sequence of 2D patches $\bm{x}_p \in\mathbb{R}^{N\times (P^2\cdot 3)}$, where $H$ and $W$ are the height and width of the image $\bm{I}_d$, $P$ is the patch size, and $N=HW/P^2$ denotes the number of patches.
Then, we flatten these patches into vectors and map them to $D$ dimensions with a linear projection.
We refer to the output projection as the patch embeddings $\bm{E}_o \in\mathbb{R}^{N\times D}$.

\smallskip
\textbf{Distortion Encoding Transformer.}
To retain positional information, positional embeddings are supplemented to the patch embeddings in transformer~\cite{vaswani2017attention}.
In our method, to avoid the simple mapping between the positional embedding and specific distortion pattern,
we arbitrarily shuffle the patch embeddings $\bm{E}_o \in\mathbb{R}^{N\times D}$ along the first dimension (token dimension).
Then, positional embeddings~\cite{vaswani2017attention} (the sine-cosine version) are added to the shuffled patch embeddings.
The resulting sequence of embedding vectors are fed into the $N_T$ transformer encoder~\cite{vaswani2017attention} layers.
Thereafter, we obtain the abstracted patch representations $\bm{E}_r \in\mathbb{R}^{N\times D}$, as shown in Fig.~\ref{framework}.

In the meantime, we build a position map $\bm{P}_o \in\mathbb{R}^{\frac{H}{P}\times\frac{W}{P}}$, in which each value denotes the distortion degree of a patch.
For clarity, we provide an example of the position map, as shown in Fig.~\ref{fig:motivation}.
Note that it can be readily obtained from the patch radius.
Besides, in Table~\ref{pct}, we show the category number $C_t$ of distortion degree, 
when dividing a $256\times256$ image with $P\times P$ patches.
Note that $C_t$ measures the level of granularity of the division on distortion degree.
Then, $\bm{P}_o$ is reshaped and shuffled in the same way as the patch embeddings.
In this way, we obtain the per-patch distortion degree annotation $\bm{P}_r \in\mathbb{R}^{N\times 1}$, corresponding to the patch representations $\bm{E}_r \in\mathbb{R}^{N\times D}$.


\smallskip
\textbf{Contrastive Loss.}\label{Contrastive}
To learn a useful distortion representation for rectification,
we first introduce a contrastive loss to supervise the output representation $\bm{E}_r \in\mathbb{R}^{N\times D}$.
The motivation of contrastive learning~\cite{chen2020simple} is to associate the query with its positive examples and push away the negative examples.
Considering that the distortion pattern of fisheye images is radially distributed (see Fig.~\ref{fig:motivation}) and the distortion degree increases with the distance from the image center,
we regard the patches with the same distortion degree as positive examples and with different degrees as negative examples, respectively.
In this way, the contrastive learning can help the network encode the local distinct distortion pattern in different image regions, regardless of their individual appearance.
Concretely, the cross-entropy loss is calculated, representing the probability of the positive examples being selected over the negatives. 
For the $i^{th} (i=1,2,\cdots,N)$ patch of the input image, the loss function is defined as follows,
\begin{equation}\label{contra_single_p}
	\ell_i=-log\frac{\sum_{j=1}^N\mathbbm{1}_{[j \in N_i]} exp(\bm{p}_i\cdot \bm{p}_j/\tau)}{\sum_{k=1}^N\mathbbm{1}_{[k \neq i]}exp(\bm{p}_i\cdot \bm{p}_k/\tau)},
\end{equation}
where $\bm{p}_i \in\mathbb{R}^{1\times D}$ is the $i^{th}$ representation in $\bm{E}_r \in\mathbb{R}^{N\times D}$, 
$N_i$ is index set of positive examples for the $i^{th}$ patch,
$\mathbbm{1}_{[f(\cdot)]} \in \{0,1\}$ represents an indicator function evaluating to 1 iff $f(\cdot)$, and $\tau$ denotes a temperature parameter.
The total contrastive loss $\mathcal{L}_{con}$ is calculated over the $N$ patches,
\begin{equation}
	\mathcal{L}_{con}=\sum_{i=1}^{N}\ell_i.
\end{equation}

\setlength{\tabcolsep}{12pt}
\begin{table}[t]
    \centering
    \vspace{0.1in}
	\begin{tabular}{ccccc}
			\hline
			$P$   &  8$\times$8   & 16$\times$16   & 32$\times$32  & 64$\times$64  \\
			
			$C_t$ &  136  & 36   & 10  &  3 \\
			\hline
	\end{tabular}
	\caption{
            The distortion degree number $C_t$ in the position map when
            dividing an $256\times256$ image with $P\times P$ patches.
            }
	\label{pct}
\end{table}

\smallskip
\textbf{Position Loss.}
As mentioned above, 
the shuffled patch representations $\bm{E}_r \in\mathbb{R}^{N\times D}$ have been labeled with $\bm{P}_r \in\mathbb{R}^{N\times 1}$ according to their distortion patterns (degrees).
To model these local distinct distortion patterns,
we further introduce a simple but effective position loss.
Specifically, given the patch representation $\bm{E}_r \in\mathbb{R}^{N\times D}$,
we apply a layer normalization~\cite{vaswani2017attention} and a linear projection to estimate their distortion degrees.
The loss is defined as a cross-entropy loss,
\begin{equation}
	\mathcal{L}_{pos} = -\sum_{n=1}^{N} \sum_{c=1}^{C_t} log\frac{exp(\bm{\hat{x}}_{n,c})}{\sum_{i=1}^{C_t}exp(\bm{\hat{x}}_{n,i})}\bm{x}_{n,c},
\end{equation}
where $\bm{x}$ and $\bm{\hat{x}}$ denote the ground truth and predicted confidence, respectively; $C_t$ is the category number of distortion degree in $\bm{P}_r$ which measures the level of granularity of the learned features.
Note that this constraint on the network also facilitates the perception of positives and negatives in the above contrastive learning.

At the pre-training stage, the overall architecture is end-to-end optimized with the following training objectives:
\begin{equation}
	\mathcal{L}_{pre} = \mathcal{L}_{con} + \mathcal{L}_{pos}.
\end{equation}

\subsection{Fine-tuning for Rectification}
After pre-training, the network is capable of extracting the fine-grained distortion features.
Next, as illustrated in Fig.~\ref{framework}, 
we initialize the rectification network with the pre-trained weights and perform fine-tuning.

Specifically, we adopt the same patch embedding module as the pre-training stage, and directly feed the embeddings $\bm{E}_o$ into $N_T$ transformer encoder layers without shuffling.
To adapt to the rectification task, we only initialize the first $N_F (N_F < N_T)$ layers with pre-trained weights, and randomly initialize the remaining layers.
The obtained patch representation $\bm{E}_r$ is then passed through two parallel modules for rectification, described below.

\begin{figure}[t]
	\begin{center}
		\includegraphics[width=0.8\linewidth]{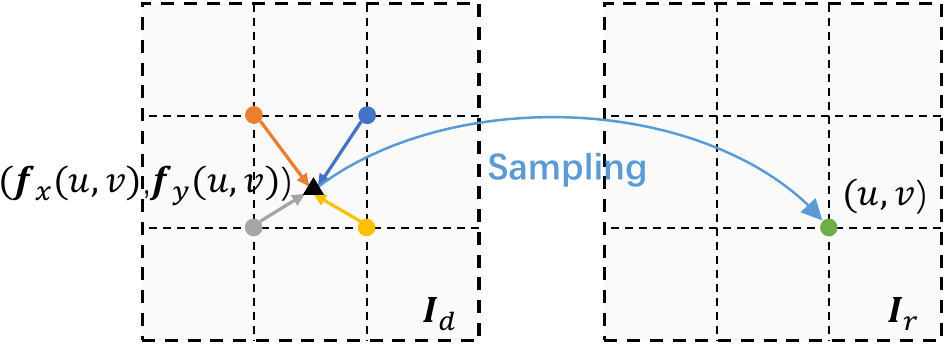}
	\end{center}
	\vspace{-0.2in}
         \caption{Visualization of the rectification process of a certain pixel on the rectified image based on warping flow. 
	}
	\label{fig:sampling}
	\vspace{-0.1in}
\end{figure}

\smallskip
\textbf{Flow Prediction Module.}
Different from the existing parameter-based and image-based methods,
we unwarp a fisheye image with a backward flow field.
Given the representation $\bm{E}_r \in\mathbb{R}^{N\times D}$,
we estimate a full scale flow $\bm{f}_b \in\mathbb{R}^{H \times W \times 2}$.
As illustrated in Fig.~\ref{fig:sampling},
with $\bm{f}_b=(\bm{f}_x,\bm{f}_y)$, the rectified image $\bm{I}_c \in\mathbb{R}^{H \times W \times 3}$ is obtained by the warping operation based on bilinear sampling as follows,
\begin{equation}\label{equ:task}
	\bm{I}_c(u,v) = \bm{I}_d(\bm{f}_x(u,v), \bm{f}_y(u,v)),
\end{equation}
where $(u,v)$ is the integer pixel coordinate in rectified image $\bm{I}_c$, and $(\bm{f}_x(u,v), \bm{f}_y(u,v))$ is the predicted decimal pixel coordinate in distorted image $\bm{I}_d$.

To this end,
we introduce a learnable upsample module.
We first reshape $\bm{E}_r$ to shape ${\frac{H}{P}\times \frac{W}{P}\times D}$.
Then, two convolutional layers (stride 1) are followed and produce a $1/P$ scale warping flow~$\bm{f}_m \in\mathbb{R}^{\frac{H}{P}\times \frac{W}{P}\times 2}$.
Next, We use other two convolutional layers to predict a $H/P\times W/P \times (P\times P\times 9)$ mask and perform softmax over the weights of the $3 \times 3$ neighborhood of each
pixel in $\bm{f}_m$.
Finally, the obtained $\frac{H}{P} \times \frac{W}{P} \times P\times P \times 2$ map is permuted and reshaped to the full resolution warping flow $\bm{f}_b \in\mathbb{R}^{H \times W \times 2}$.

\smallskip
\textbf{Boundary Refinement Module.}\label{BRM}
We find that the obtained $\bm{I}_c$ struggles with the serrated boundaries and noisy backgrounds (see Fig.~\ref{mask_aba}).
To further improve the image quality,
we propose a boundary refinement module (BRM).

Similar to the flow prediction module, 
BRM also estimates the weights of neighbor pixels for upsampling.
Then,
we obtain a confidence map $\bm{M} \in\mathbb{R}^{H \times W\times 1}$, in which each value denotes the confidence whether a pixel belongs to the rectified region.
$\bm{M}$ is further binarized with a threshold $\sigma$ to obtain the binary boundary refinement mask $\bm{M}_b \in\mathbb{R}^{H \times W\times 1}$,
as illustrated in Fig.~\ref{framework}.
The final rectified image $\bm{I}_r$ can be obtained by multiplying $\bm{I}_c$ with $\bm{M}_b$.

\smallskip
\textbf{Training Loss.}
At the fine-tuning stage,
the model is end-to-end optimized with the following training objectives,
\begin{equation}
	\mathcal{L}_{ft} = \mathcal{L}_{flow} + \mathcal{L}_{mask}.
\end{equation}

$\mathcal{L}_{flow}$ is the $L_1$ distance between the predicted warping flow $\bm{f}_b$ and its given ground truth $\bm{f}_{gt}$ as follows,
\begin{equation}
	\mathcal{L}_{flow} = \left \| \bm{f}_b \times \bm{M}_{gt}  - \bm{f}_{gt} \right \|_1,
\end{equation}
where $\bm{M}_{gt}$ denotes the ground truth valid foreground region of the rectified image.

$\mathcal{L}_{mask}$ is the cross-entropy loss between the predicted boundary refinement confidence map $\bm{M}$ and its given ground truth $\bm{M}_{gt}$ as follows,
\begin{equation}
	\mathcal{L}_{mask} = -\sum_{i=1}^{N_d}\left[\bm{y}_i\log(\hat{\bm{y}_i})+(1-\bm{y}_i)\log(1-\hat{\bm{y}_i})\right],
\end{equation}
where $N_d=H\times W$ is the number of pixels in image $\bm{I}_d$, $\bm{y}_i \in \{0, 1\}$ and $\hat{\bm{y}_i} \in [0, 1]$ denote the ground-truth and the predicted confidence value in $\bm{M}_{gt}$ and $\bm{M}$, respectively.

\normalsize
\section{Experiments}
\subsection{Dataset and Metrics}
\label{datasettext}
We train and evaluate our method according to the conventions in the literature.
Following method~\cite{rong2016radial,bogdan2018deepcalib,yang2021progressively,liao2021multi,liao2019dr},
we establish a dataset using original images from the Place365 dataset~\cite{zhou2017places}.
We use the polynomial model (Eq.~\eqref{distortion_model}) to generate the dataset
and the first four distortion parameters are used ($N_k = 4$), which is enough to fit most of the real-world application scenarios.
The synthetic dataset contains multiple annotations for each distorted fisheye image,
including distortion-free image, warping flow, and distortion parameters.
We synthesize 100k images for training and 5k images for evaluation, respectively.

In fisheye image rectification,
Peak Signal to Noise Ratio (PSNR) and Structural Similarity (SSIM) are the two most popular
evaluation metrics in existing works.
Given a rectified image and a reference distortion-free one, 
PSNR measures the detailed quality differences while SSIM assesses their structural differences.
Besides, we use Frechet Inception Distance (FID) to evaluate the distribution difference.

\setlength{\tabcolsep}{9pt}
\small
\begin{table*}[t]
    \small
	\renewcommand\arraystretch{1}
	\centering
	\subfigcapskip=1pt
	\subtable[\textbf{Pre-training.} Joint pre-training obtains remarkable performance gain.]{
		\label{pre-training}
		\begin{tabular}{cccc}
			$\mathcal{L}_{con}$ & $\mathcal{L}_{pos}$ & PSNR & SSIM \\
			\hline
			& & 20.33 & 0.82 \\
			$\checkmark$ & & 20.70 &  0.83 \\
			& $\checkmark$ & 21.30 & 0.84 \\
			$\checkmark$ & $\checkmark$ & \cellcolor{gray!20}\textbf{22.47} & \cellcolor{gray!20}\textbf{0.86} \\
    \end{tabular}}
	\quad
    \subtable[\textbf{Fine-tuning way.} Both fine-tuning strategies work and training from-scratch is better.]{
    	\label{Fine-tuning way}
    	\begin{tabular}{ccc}
    		settings & PSNR & SSIM \\
    		\hline
    		w/o pre-training & 20.33 & 0.82 \\
            \textbf{\emph{w/ pre-training}}: \\
    		frozen encoder  & 21.12 & 0.85 \\
    		unfrozen encoder & \cellcolor{gray!20}\textbf{22.47} & \cellcolor{gray!20}\textbf{0.86} \\
	\end{tabular}} 
	\quad
	\subtable[\textbf{Shuffle.} Without the shuffle, the performance drops to the baseline level.]{
		\label{shuffle}
		\begin{tabular}{ccc}
			settings & PSNR & SSIM \\
			\hline
			w/o shuffle & 20.28 & 0.82 \\
			w/ shuffle & \cellcolor{gray!20}\textbf{22.47} & \cellcolor{gray!20}\textbf{0.86} \\
	\end{tabular}}\vskip 4pt
	\quad
	\subtable[\textbf{Prediction target.} Compared with distortion parameter and image, warping flow is the most effective target.]{
		\label{reconstruction target}
		\begin{tabular}{ccc}
			targets & PSNR & SSIM \\
			\hline
			parameter & 21.09 & 0.82 \\
			image & 20.50 & 0.69 \\
			flow & \cellcolor{gray!20}\textbf{22.47} & \cellcolor{gray!20}\textbf{0.86} \\
	\end{tabular}} 
    \quad
	\subtable[\textbf{Patch size.} Reducing patch size works better. To strike a balance between accuracy and  efficiency, we choose 16$\times$16 by default.]{
		\label{patch size}
		\begin{tabular}{cccc}
			patch & PSNR & SSIM & FPS \\
			\hline
			8 & \textbf{25.82} & \textbf{0.90} & 14.93 \\
			16 & \cellcolor{gray!20}22.47 & \cellcolor{gray!20}0.86 & \cellcolor{gray!20}\textbf{15.51} \\
			32 & 20.23 & 0.81 & 14.64 \\
			64 & 18.65 & 0.74 & 14.19 \\
	\end{tabular}}
    \quad
	\subtable[\textbf{Architecture setting.} Several different architectures are set up. Our scheme performs better.]{
		\label{settings}
		\begin{tabular}{ccc}
			settings & PSNR & SSIM \\
			\hline
			w/o BRM & 18.01 & 0.68 \\
			w/o FPM & 22.11 & 0.85 \\
			full & \cellcolor{gray!20}\textbf{22.47} & \cellcolor{gray!20}\textbf{0.86} \\
	\end{tabular}}
    \vspace{-0.12in}
	\caption{Ablation experiments of our method. 
    The default setting is: the patch size is $16\times 16$ and we import 8 pre-trained layers for fine-tuning.
    Default settings are all marked in \colorbox{gray!20}{gray}.}
	\label{ablation}
     \vspace{-0.1in}
\end{table*}

\normalsize
\subsection{Implementation}

\textbf{Model Settings.} 
The image size $(H, W)$ is (256, 256).
The latent vector size $D$ in transformer encoder layer~\cite{vaswani2017attention} is $256$ and the total layer number $N_T$ is $10$.
The threshold $\sigma$ for binarizing the confidence map in the boundary refinement module is empirically set as 0.5.

\smallskip
\textbf{Training Details.} We use 100k images for pre-training and 50k images for fine-tuning. For contrastive learning, we set the temperature $\tau = 0.07$ (Eq.~\eqref{contra_single_p}). We utilize Adam~\cite{kingma2014adam} as the optimizer and one-cycle policy~\cite{smith2019super,teed2020raft} with a maximum learning rate of $10^{-4}$.
Both stages are trained for 65 epochs with a batch size of 64.
Two NVIDIA GeForce RTX 2080Ti GPUs are used to train the network.

\subsection{Ablation Study}
We conduct ablations to verify the effectiveness of the core settings and components in our method.

\begin{figure}[t]
	\centering
	\subfigure[flow loss]{
		\label{flow_loss}
		\includegraphics[width=0.47\linewidth]{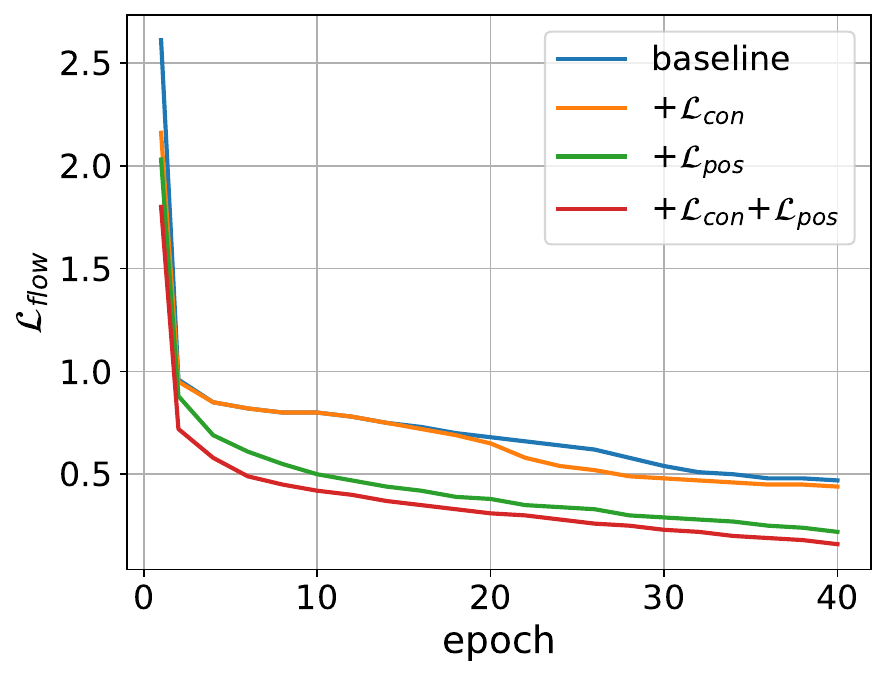}}
	\subfigure[mask loss]{
		\label{mask_loss}
		\includegraphics[width=0.48\linewidth]{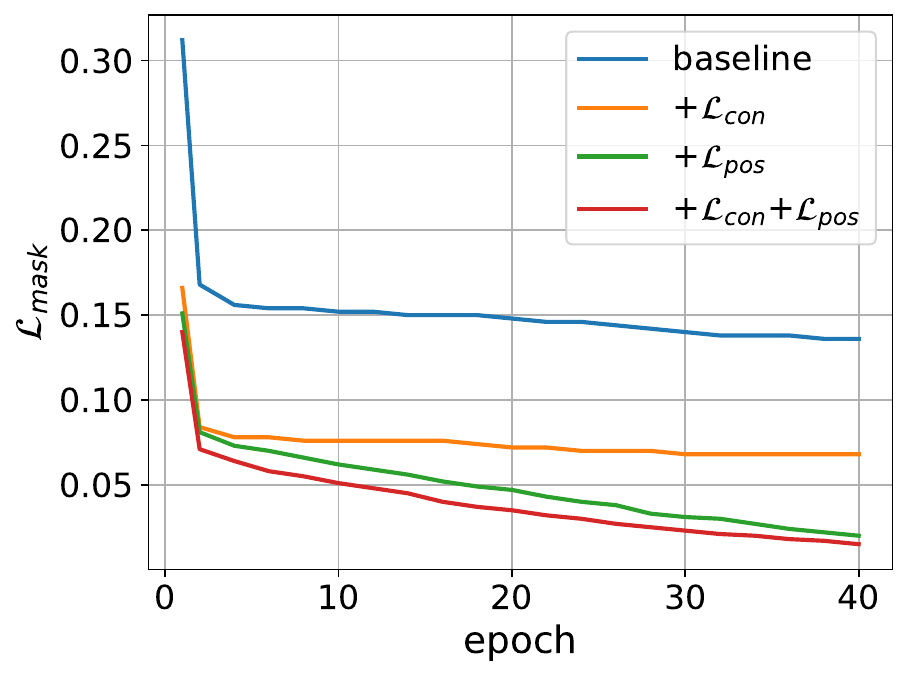}}
	\vspace{-0.1in}
	\caption{
            Training loss plots during the fine-tuning stage. 
            With the distortion representation pre-training, 
            the training errors of the rectification ($\mathcal{L}_{flow}$ and $\mathcal{L}_{mask}$) are reduced.
	}
	\label{loss}
    \vspace{-0.1in}
\end{figure}

\smallskip
\textbf{Impact of Self-supervised Pre-training.}
The key idea of this work is the self-supervised distortion representation learning for fisheye images.
Table~\ref{ablation} (a) shows the studies on the proposed pretext task.
Without any pre-training, we can still provide a strong baseline, with a PSNR of 20.33 and a SSIM of 0.82.
This can be attributed to the powerful representative ability of transformer~\cite{vaswani2017attention}.
Then, compared with the baseline,
the performance improves by pre-training with an independent contrastive loss $\mathcal{L}_{con}$ or position loss $\mathcal{L}_{pos}$. 
Furthermore, the unified pre-training boosts the rectification performance significantly.
These results verify the effectiveness of our self-supervised representation learning strategy.
Note that the two pre-text tasks promote each other. 
In Fig.~\ref{loss}, we provide the training loss plots of fine-tuning under the setting w/ and w/o the pre-training.
From the training perspective, we can observe that our pre-training effectively reduces the losses of the baseline.

\smallskip
\textbf{Impact of Fine-tuning Way.}
During the fine-tuning stage, we train the model from scratch.
An alternative way is to freeze the pre-trained encoder and only update the rest weights of the model.
As shown in Table~\ref{ablation} (b),
both fine-tuning ways produce better performance than the baseline (w/o pre-training),
illustrating the effectiveness of our learned representations and our default fine-tuning strategy.

\smallskip
\textbf{Impact of Shuffle Manipulation.}
Shuffle strategy is important for our method.
During the pre-training stage,
we arbitrarily shuffle the patches, to avoid the directed association between the fixed positional embedding and the specific distortion pattern/degree.
As shown in Table~\ref{ablation} (c), without the shuffle strategy, the performance degrades to the baseline level in Table~\ref{ablation} (a),
which means that the pre-training does not learn any useful representation for the downstream rectification.

\begin{figure}[t]
	\centering
	\includegraphics[width=0.98\columnwidth]{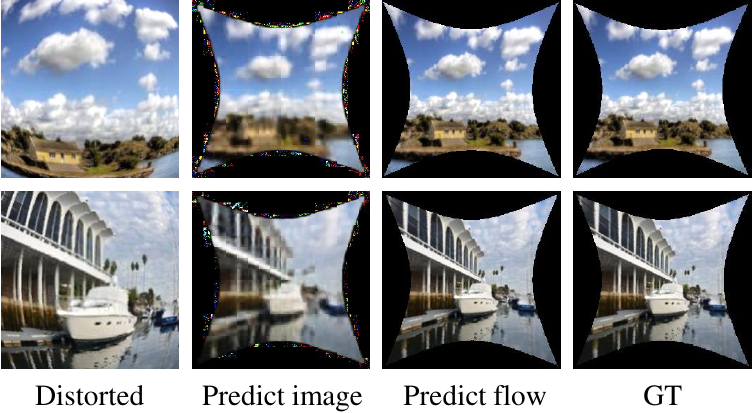}
	\caption{Examples to illustrate the impact of the reconstruction target. For our method, predicting flow to unwarp the distorted images preserves more image details.}
	\label{predtarget}
        \vspace{-0.05in}
\end{figure}

\begin{figure}[t]
	\centering
	\includegraphics[width=0.78\columnwidth]{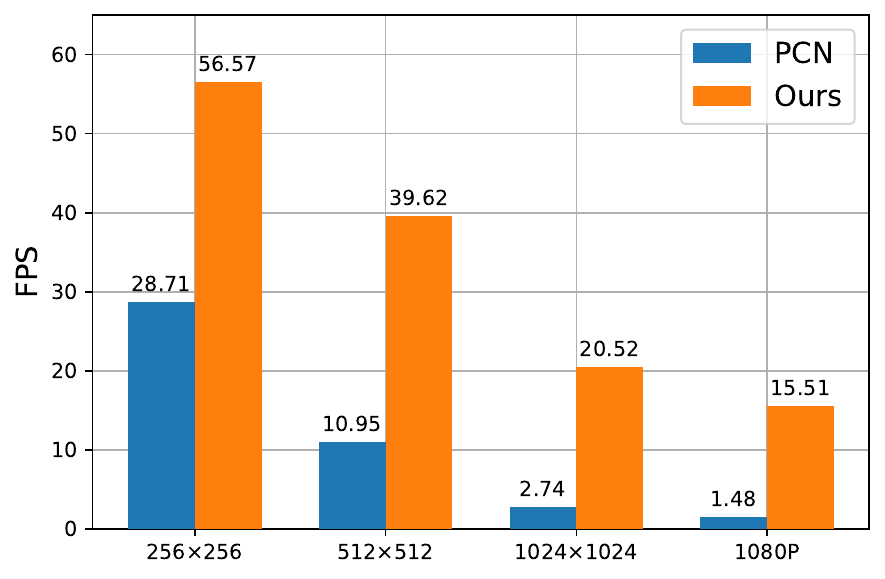}
	\vspace{-0.05in}
	\caption{Running efficiency comparison between our SimFIR and the state-of-the-art     PCN~\cite{yang2021progressively} on processing different resolution fisheye images.}
	\label{Efficiency}
        \vspace{-0.1in}
\end{figure}

\smallskip
\textbf{Impact of Reconstruction Target.}
Our SimFIR estimates a warping flow~\cite{feng2022geometric,feng2023deep} to resample the pixels from distorted images for rectification.
In contrast,
other pipelines predict the distortion parameters~\cite{rong2016radial,bogdan2018deepcalib,yin2018fisheyerecnet,xue2019learning} or regress the rectified images directly~\cite{liao2020model,yang2021progressively,liao2019dr}.
We implement the above two pipelines into our method with the following experiments: 1) meaning the output representations $\bm{E}_r \in\mathbb{R}^{N\times D}$ into a vector and projecting it to regress the distortion parameters, and 2) reshaping the output $\bm{E}_r$ to form a reconstructed image, following MAE~\cite{he2022masked}.
As shown in Table~\ref{ablation} (d), the performance drops with these two targets.
This is because for distortion rectification, predicting the deformation instead of the target image itself is simpler~\cite{li2019document,feng2021docscanner}, and can preserve more image details (see Fig.~\ref{predtarget}).
The network can focus on the spatial transformation between two images, 
without needing to restore the texture details.
Besides, it is difficult to extract complex nonlinear model parameters in different intervals~\cite{yang2021progressively} while the pixel-wise warping flow provides a more explicit and direct expression for the distortion model.

\begin{figure}[t]
	\centering
	\includegraphics[width=0.98\columnwidth]{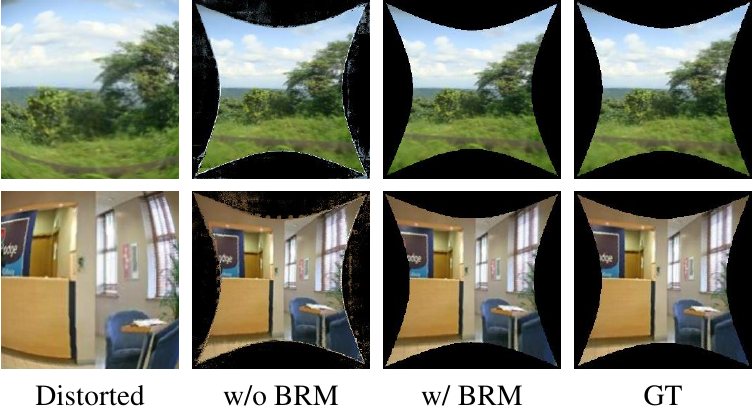}
	\caption{Examples to illustrate the impact of the boundary refinement module (abbreviated as “BRM”). With the help of BRM, the boundary regions are well rectified.}
	\label{mask_aba}
\end{figure}

\smallskip
\textbf{Impact of Patch Size.}
Table~\ref{ablation} (e) studies the impact of patch size.
We can observe that a smaller patch steadily produces a higher performance.
The reason is that, as shown in Table~\ref{pct}, a smaller patch setting increases the subdivision of distortion patterns, 
which helps to extract more fine-grained distortion representations.
However, a smaller patch will also introduce more patches when dividing an input image. 
To balance the computational cost, 
we choose a $16 \times 16$ patch size as the default setting in our final model. 

\setlength{\tabcolsep}{11pt}
\begin{table}[t]
\renewcommand\arraystretch{1.05}
\centering
        \small
	\begin{tabular}{c|cccc}
	     \hline
		Method & PSNR $\uparrow$ & SSIM $\uparrow$ & FID $\downarrow$  \\
        \hline
		SC~\cite{santana2016iterative} & 10.14 & 0.32 & 229.60 \\
		DeepCalib~\cite{bogdan2018deepcalib} & 13.28 & 0.51 & 136.73 \\
		DR-GAN~\cite{liao2019dr} & 18.66 & 0.70 & 74.56    \\
		DDM~\cite{liao2020model} & 17.51 & 0.66 & 114.66    \\
		MLC~\cite{liao2021multi} & 19.33 & 0.73 & 55.97  \\
		PCN~\cite{yang2021progressively} & 20.53 & 0.76 & 30.64 \\
		Ours & \textbf{22.47} & \textbf{0.86} & \textbf{17.23}  \\
         \hline
	\end{tabular}
	\caption{
	Comparison between the state-of-the-art methods and the proposed method on the test set.
	``$\uparrow$'' indicates the higher the better and ``$\downarrow$'' means the opposite.}
	\label{Quantitative}
      \vspace{-0.1in}
\end{table}

\begin{figure*}[t]
	\centering
	\includegraphics[width=2\columnwidth]{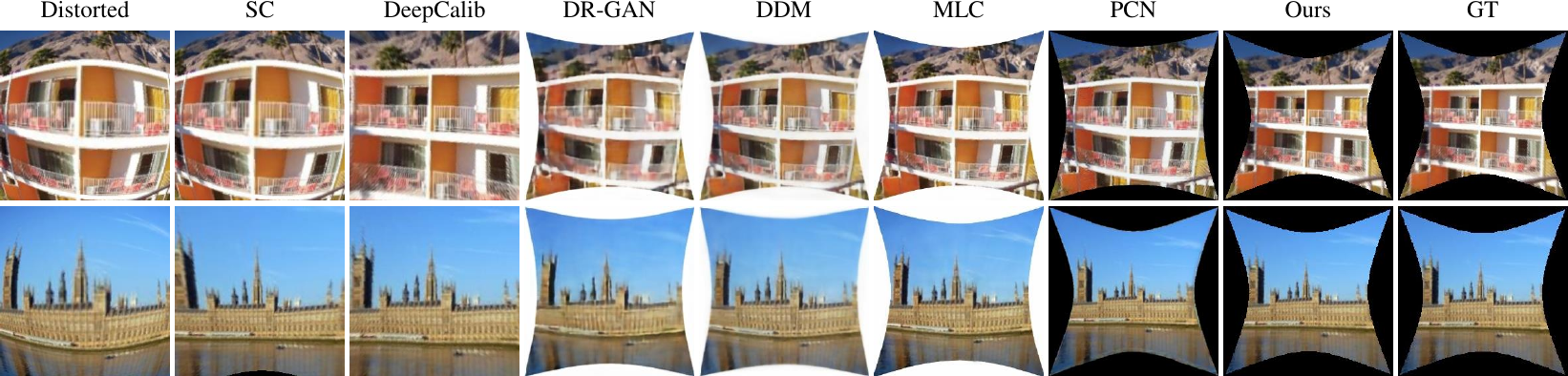}
     \vspace{-0.04in}
	\caption{Qualitative results on the synthesized test set. For each comparison, we show the distorted image, the rectified results of SC~\cite{santana2016iterative}, DeepCalib~\cite{bogdan2018deepcalib}, DR-GAN~\cite{liao2019dr}, DDM~\cite{liao2020model}, MLC~\cite{liao2021multi}, PCN~\cite{yang2021progressively}, our method, and ground truth.
	}
	\label{qua_synthesized}
\end{figure*}

\begin{figure*}[h]
	\centering
	\includegraphics[width=2\columnwidth]{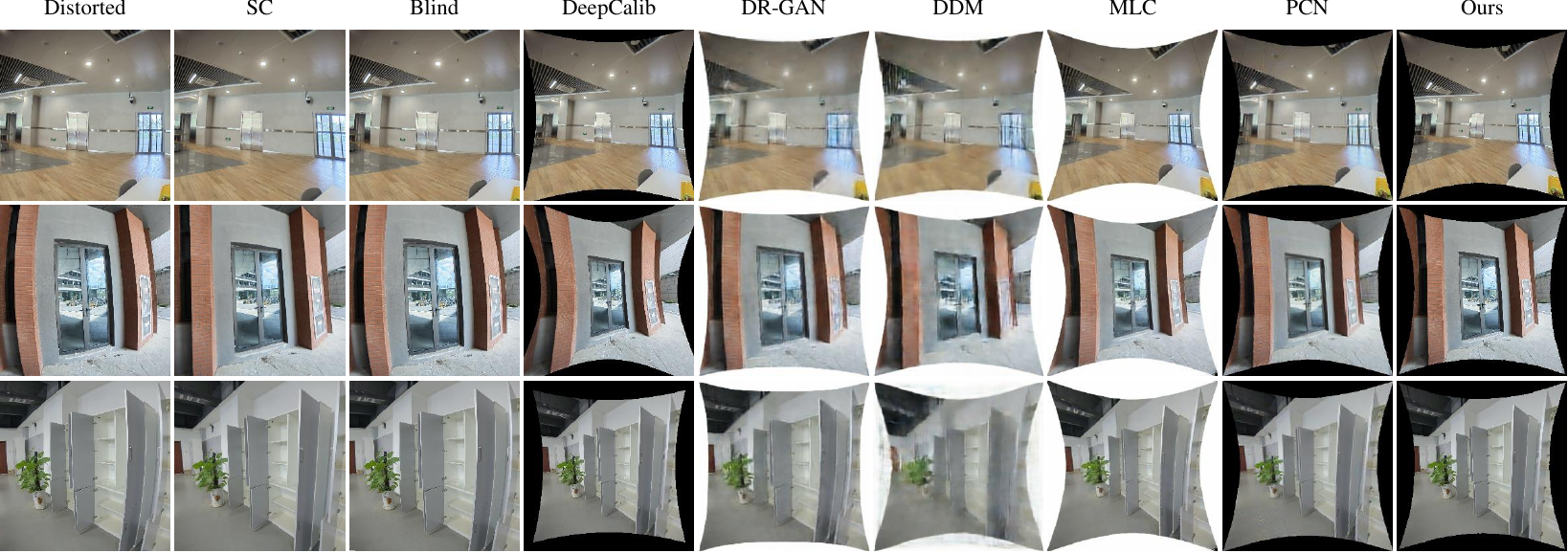}
	\vspace{-0.04in}
	\caption{Qualitative results on real-world fisheye images. 
	For each comparison, we show the distorted image, the rectified results of SC~\cite{santana2016iterative}, Blind~\cite{li2019blind}, DeepCalib~\cite{bogdan2018deepcalib}, DR-GAN~\cite{liao2019dr}, DDM~\cite{liao2020model}, MLC~\cite{liao2021multi}, PCN~\cite{yang2021progressively}, and our method.
	}
	\label{real}
\end{figure*}

\smallskip
\textbf{Impact of Boundary Refinement Module.}
In our SimFIR, the BRM is used to smooth the serrated boundaries and remove the noisy backgrounds of the coarse rectified image $\bm{I}_c$ (see Sec.~\ref{BRM}).
As shown in Table~\ref{ablation} (f), with the proposed BRM, performances are remarkably improved.
Moreover, we provide two examples to visualize the improvement from the BRM in Fig.~\ref{mask_aba}.
As we can see, the boundary regions are well rectified with the help of BRM.

\smallskip
\textbf{Impact of Flow Prediction Module.}
We compare the bilinear upsampling with our learnable upsampling used for FPM.
Table~\ref{ablation} (f) shows that our learnable FPM works better than the module based on the bilinear upsampling.
This is likely because the coarse bilinear upsampling operation can not recover the small deformations.

\subsection{Comparison with State-of-the-art Methods}
\textbf{Quantitative Comparison.}
In this section, we compare our SimFIR with the recent state-of-the-art rectification methods, including SC~\cite{santana2016iterative}, DeepCalib~\cite{bogdan2018deepcalib}, DR-GAN~\cite{liao2019dr}, 
DDM~\cite{liao2020model}, MLC~\cite{liao2021multi}, and PCN~\cite{yang2021progressively}.
The quantitative results are shown in Table~\ref{Quantitative}.
As we can see, our method achieves state-of-the-art performance on all metrics.
Note that for SC~\cite{santana2016iterative}, DeepCalib~\cite{bogdan2018deepcalib}, and PCN~\cite{yang2021progressively}, we calculate the results based on their released codes.
For DR-GAN~\cite{liao2019dr}, DDM~\cite{liao2020model}, and MLC~\cite{liao2021multi}, we obtain the rectified test set from the authors.

\smallskip
\textbf{Qualitative Comparison.}
To better demonstrate the effectiveness of our proposed method, 
we further conduct qualitative comparisons with existing methods~\cite{santana2016iterative, bogdan2018deepcalib, liao2019dr, liao2020model, liao2021multi, yang2021progressively} on the synthetic test set.
As present in Fig.~\ref{qua_synthesized}, the rectified images of our method show less distortion.

\smallskip
\textbf{Application to Real-world Fisheye Images.}
Additional experiments are conducted on real-world fisheye images.
In this evaluation,
we capture the images with several fisheye lenses of different FoV.
As shown in Fig.~\ref{real}, although there exists a gap between synthetic and real-world data, 
our method still shows strong generalization ability.
Compared with SC~\cite{santana2016iterative}, our method guarantees the integrity of the recovered content.
Compared with Blind~\cite{li2019blind}, DeepCalib~\cite{bogdan2018deepcalib}, DR-GAN~\cite{liao2019dr}, 
DDM~\cite{liao2020model}, MLC~\cite{liao2021multi}, and PCN~\cite{yang2021progressively},
our rectified images show less distortion.
Note that in Fig.~\ref{real}, the resolution of the input distorted fisheye images is much higher (up to 2454 $\times$ 2454) than that in the test set (256 $\times 256$).
Compared with DR-GAN~\cite{liao2019dr}, DDM~\cite{liao2020model}, and PCN~\cite{yang2021progressively}, our method preserves more texture details.

\smallskip
\textbf{Efficiency Comparison.}
Next, we compare the running efficiency of our method and the state-of-the-art method PCN~\cite{yang2021progressively} on processing images with various resolutions.
As shown in Fig.~\ref{Efficiency},
on different resolutions, our method shows higher efficiency.
The reason is that PCN~\cite{yang2021progressively} directly regresses the rectified image with a fully convolutional network~\cite{ronneberger2015u}, like DR-GAN~\cite{liao2019dr}, and DDM~\cite{liao2020model}.
In contrast, our SimFIR infers a warping flow at the fixed resolution ($256 \times 256$) and interpolates it to the original high resolution same as the input image for unwarping.
Such a strategy ensures efficiency and does not demand any extra training at a certain scale like image-based methods. 


\section{Conclusion}
In this work, we present an effective self-supervised representation learning paradigm for fisheye images.
We associate different image regions with their specific distortion patterns and design a distortion-aware pretext task for their learning. 
With the pre-training, the network extracts the fine-grained distortion representation.
Extensive experiments are conducted on our proposed dataset, and the transfer performance in the downstream rectification task verifies the effectiveness of the learned representations.


\smallskip
\smallskip
\noindent
\textbf{{Acknowledgements}}
This work was supported by NSFC under Contract 61836011 and 62021001. It was also supported by the GPU cluster built by MCC Lab of Information Science and Technology Institution, USTC, and the Supercomputing Center of the USTC.

{\small
\bibliographystyle{ieee_fullname}
\bibliography{egbib}
}

\end{document}